%% file: main.tex
\documentclass[runningheads]{llncs}

\usepackage[T1]{fontenc}
\usepackage[utf8]{inputenc}
\usepackage{graphicx}
\usepackage{booktabs}
\usepackage{array}
\usepackage{multirow}
\usepackage{amsmath,amssymb}
\usepackage{microtype}
\usepackage{xcolor}
\usepackage{tikz}
\usepackage{hyperref}
\usepackage[capitalise,noabbrev]{cleveref}
\usepackage{caption}
\usepackage{subcaption}
\usepackage{enumitem}
\usepackage{placeins}

\hypersetup{
  colorlinks=true,
  linkcolor=blue!60!black,
  citecolor=teal!60!black,
  urlcolor=blue!60!black,
}

\definecolor{orcidgreen}{HTML}{A6CE39}
\newcommand{\orcidbadge}[1]{%
  \href{https://orcid.org/#1}{%
    \raisebox{0.35ex}{%
      \tikz[baseline=-0.55ex]{%
        \node[circle,fill=orcidgreen,inner sep=0pt,minimum size=1.65ex]
          {\color{white}\sffamily\fontsize{4.2}{4.2}\selectfont iD};}}}}
\renewcommand{\orcidID}[1]{\unskip\,\orcidbadge{#1}}

\newcommand{\softbias}{\textsc{soft\_\allowbreak bias}}
\newcommand{\softpos}{\textsc{soft\_\allowbreak pos\_\allowbreak only}}
\newcommand{\softcontent}{\textsc{soft\_\allowbreak content\_\allowbreak only}}

\newcommand{\unforced}{\texttt{unforced}}
\newcommand{\forcedzero}{\texttt{forced\_\allowbreak zero}}
\newcommand{\forcedconstHalf}{\texttt{forced\_\allowbreak const0.5}}
\newcommand{\linearHalfZero}{\texttt{linear0.5\allowbreak\ensuremath{\rightarrow}0}}
\newcommand{\fadeStartSixHundredZero}{\texttt{fade600\allowbreak\ensuremath{\rightarrow}0}}
\newcommand{\Bpos}{B^{\text{pos}}}
\newcommand{\Bcontent}{B^{\text{content}}}

\begin{document}

\title{\fontsize{13pt}{15.5pt}\selectfont Training Trajectories Determine Circuit Removability\\[-0.1em]
in Annealable Soft-Prior Transformers}
\titlerunning{Training Trajectories Determine Circuit Removability}

\author{Zonglin Yang\orcidID{0009-0000-5732-5699}\inst{1}\thanks{Corresponding author.}\and
Ziming Zhao\inst{2}\and
Wei Tang\inst{2}\and
Xunyu Jiang\inst{2}\and
Yihong Liu\inst{2}\and
Tailin Chen\inst{2}\and
Zifu Yu\inst{3}\and
Jiayu Liu\inst{1}}
\authorrunning{Z. Yang et al.}
\institute{Criminal Science and Technology, Guangdong Police College, China\\
\email{3258244847@qq.com; 3960432092@qq.com}
\and
Cybersecurity and Law Enforcement, Guangdong Police College, China\\
\email{765979068@qq.com; 3311739267@qq.com; 1581367121@qq.com}\\
\email{zzm1030082097@gmail.com; 3278324576@qq.com}
\and
Traffic Management Engineering, Guangdong Police College, China\\
\email{yuzifu@yuzifu.top}}

\maketitle

\input{sections/abstract}
\input{sections/introduction}
\input{sections/related_work}
\input{sections/method}
\input{sections/experimental_setup}
\input{sections/main_results}
\input{sections/causal_analysis}
\input{sections/training_path_boundary}
\input{sections/cross_task}
\input{sections/mechanistic}
\input{sections/limitations}

\input{sections/conclusion}
\section*{Funding}

This work was supported by the Guangdong S\&T Programme through the project
``Key Technologies and Applications for Proactive Monitoring and Early Warning of
Public Security Risks Based on Multimodal Large Models''
(Project No.~2026B0101100001).

\section*{Conflict of Interest}

The authors declare that they have no conflicts of interest.

\FloatBarrier
\bibliographystyle{splncs04}
\bibliography{references}

\end{document}

%% file: sections/abstract.tex
\begin{abstract}
Soft positional priors can help small Transformers learn retrieval
circuits, but it is unclear whether the resulting circuits remain
functional once the prior is removed. We test this with an annealable
soft-prior Transformer whose attention biases can be learned, faded, or
zeroed during training and evaluation. On associative recall, unforced
models perform well with the prior active ($0.772 \pm 0.020$) but collapse
at zero gate ($0.095 \pm 0.009$). Smooth fade-to-zero training preserves
high zero-gate accuracy ($0.734 \pm 0.028$), whereas forced-zero training,
hard switching, and post hoc continuation fail to recover the same effect.
The pattern also appears on Markov induction. Linear regression ICL provides
a boundary case because zero-gate training can learn that task directly.
Mechanistic traces show that circuit consolidation occurs after the gate
reaches zero, even though the responsible heads vary across seeds. These
results suggest that circuit removability in small discrete retrieval tasks
depends on the training trajectory, not just the final architecture.

\keywords{In-context learning \and Induction heads \and Positional encoding
\and Training dynamics \and Mechanistic interpretability \and Annealing.}
\end{abstract}

%% file: sections/introduction.tex
\section{Introduction}
\label{sec:intro}

In small Transformers trained on synthetic in-context learning (ICL) tasks,
a single inductive bias can determine whether a retrieval circuit forms.
Relative position biases, local priors, and specialized positional encodings
can make recall of values by key and induction much easier~\cite{press2022train,shaw2018self,li2024functional,chi2022kerple,d2021convit}. Standard evaluation leaves these priors active at inference,
so it cannot distinguish a circuit that works without the prior from one
that depends on the prior permanently.

We ask whether the prior can be removed after it helps a circuit form.
Removability distinguishes two mechanisms under the same architecture: the
model either learns an internal attention pattern that implements retrieval
or continues to rely on an external bias term. Removing the bias at
evaluation tests which mechanism the model uses.

We add gated position and content biases to each attention head. The gate
value can be learned freely, clamped, or scheduled during
training. On associative recall, freely trained models achieve high normal
accuracy but fail at zero gates. A fade-to-zero trajectory, which holds the
gate at $0.5$, smoothly reduces it to $0$, and then continues training with
the prior absent, preserves high zero-gate accuracy. Neither a warm start nor
extra zero-gate training explains the result: abrupt switches and post hoc
continuation from an unforced checkpoint both fall far short.

Several smooth training paths work, but abrupt removal does not.
Mechanistic traces show that the heads
that carry retrieval differ across seeds, while the timing of consolidation
is stable and occurs after the gate is already zero. The same pattern holds
on Markov induction, whereas linear regression ICL behaves differently:
forced-zero training solves the task, so it can form a circuit without the
prior, unlike the discrete retrieval tasks.

\begin{figure}[t]
\centering
\includegraphics[width=\linewidth]{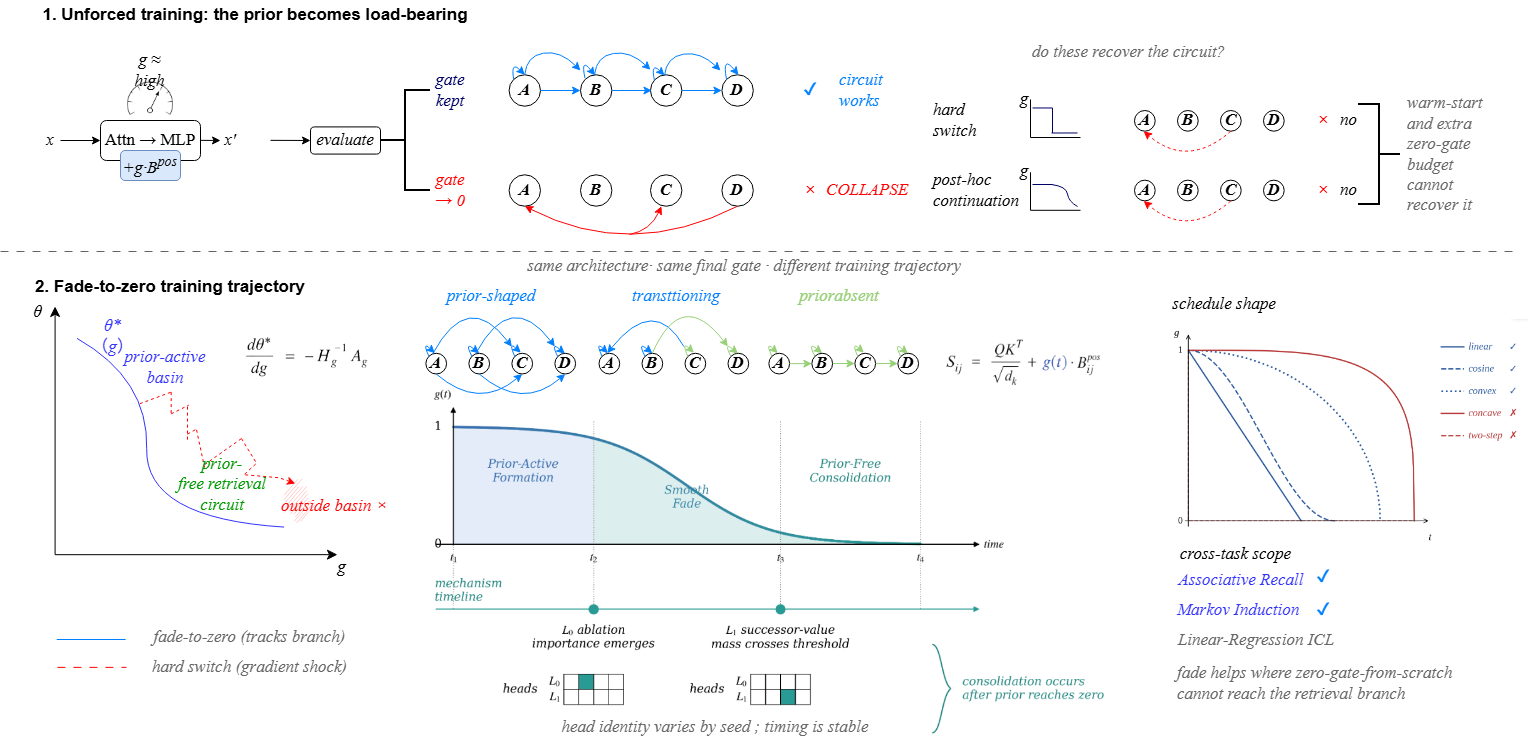}
\caption{Conceptual overview. Under unforced training, the positional prior
becomes necessary: the model works with an active gate but collapses when
the gate is zeroed. With fade-to-zero training, the model first forms a
retrieval circuit shaped by the prior, follows the gate as it decreases,
and consolidates a prior-free circuit. Hard switches and post hoc
continuation test whether a warm start or extra zero-gate training can
recover the same circuit.}
\label{fig:conceptual_overview}
\end{figure}

\newpage
\paragraph{Contributions.}
\begin{itemize}
\item We introduce an annealable soft-prior Transformer and evaluation
protocol that distinguishes prior-active performance from prior-free circuit
removability.
\item We show that fade-to-zero training preserves zero-gate retrieval on
associative recall and Markov induction, while unforced training,
forced-zero training, hard switching, post hoc continuation, prior dropout,
and a dual condition loss do not reproduce the same effect. Schedule sweeps
and three-seed mechanism traces further show that smooth removal produces
stable consolidation timing despite variation in head identity.
\item We identify a boundary case in linear regression ICL, where
removability failure exists for unforced training but a zero-gate model can
learn the task directly.
\end{itemize}

%% file: sections/related_work.tex
\section{Related Work}
\label{sec:related}

\paragraph{Position and attention priors.}
Transformers depend strongly on how position is represented
inside attention~\cite{vaswani2017attention,shaw2018self,press2022train,su2024roformer,kazemnejad2023impact}. Recent methods such as
\textsc{Kerple}, \textsc{FIRE}, \textsc{CoPE}, and \textsc{DAPE} design
position functions that improve length extrapolation or adapt position
signals to the data~\cite{chi2022kerple,li2024functional,golovneva2024contextual,zheng2024dape}. GPSA/ConViT adds a soft convolutional
locality prior to vision Transformers~\cite{d2021convit}. These
methods usually treat the prior as part of the final architecture. We
instead ask whether a prior that helps during training remains necessary
after the circuit forms.

\paragraph{In-context learning and circuits.}
Linear regression ICL shows that Transformers can implement simple learning
algorithms in context~\cite{garg2022can,akyurek2022learning}. Discrete retrieval
tasks expose induction heads and successor circuits that can be measured
directly~\cite{olsson2022context,nanda2023progress}. Work on circuit
interpretability and automated circuit discovery studies how trained
attention heads implement specific computations~\cite{wang2022interpretability,conmy2023towards}. These tasks expose the relevant circuit clearly
enough to test whether the prior remains causally necessary.

\paragraph{Continuation paths and soft targets.}
The fade-to-zero schedule follows the logic of homotopy and continuation
methods, which solve a hard problem by tracking solutions from an easier
nearby problem as a control parameter changes~\cite{allgower2012numerical,mobahi2015link}. Curriculum learning also changes the optimization
path by presenting easier training conditions first~\cite{bengio2009curriculum}. Knowledge distillation uses soft targets to shape gradients before
the final hard prediction objective is reached~\cite{hinton2015distilling}.
Our schedule differs in where the auxiliary signal enters: it is an
prior added to the attention logits, not an output target or data curriculum, and we test
whether the induced circuit survives after the signal is removed.

\paragraph{Removable and reparameterized structure.}
Stochastic depth, LayerDrop, movement pruning, and lottery ticket methods
study networks that remain useful after removing layers, weights, or
other components~\cite{huang2016deep,fan2019reducing,sanh2020movement,frankle2018lottery}. Structural reparameterization methods such as
RepVGG train with richer branches and convert them into a simpler inference
graph~\cite{ding2021repvgg}. Early convolutional stems in vision
Transformers also show that an inductive bias can affect optimization even
when later representations become less explicitly convolutional~\cite{xiao2021early}. We hold the final architecture fixed, change only the gate
trajectory of an attention prior, and test whether the learned retrieval
circuit remains functional at zero gate.

%% file: sections/method.tex
\section{Method}
\label{sec:method}

\subsection{Annealable Soft-Prior Attention}
\label{sec:method_attention}

For each attention head, let $Q,K,V \in \mathbb{R}^{T \times d}$ be the
query, key, and value matrices. We add two gated bias terms before the
softmax:
\begin{equation}
S_{ij}
= \frac{Q_iK_j^\top}{\sqrt{d}}
+ g_\text{pos}(t)\Bpos_{ij}
+ g_\text{content}(t)\Bcontent_{ij}.
\label{eq:soft_prior_score}
\end{equation}
The positional prior $\Bpos$ is a learnable table indexed by clipped relative position.
The content prior is the embedding inner product
$\Bcontent_{ij}=E_i^\top E_j/\sqrt{d_e}$, computed from learned input token
embeddings rather than contextual states. Each gate is scalar per head and
lies in $[0,1]$.

\paragraph{Gate modes.}
Each head owns a gate logit $\ell$ with raw value $\sigma(\ell)$. Training
uses one of three modes:
\begin{description}[leftmargin=1.2em,style=nextline]
\item[\textsc{Free}.] $g(t)=\sigma(\ell)$ and the logit receives gradients.
\item[\textsc{Forced}.] $g(t)=c$ for a fixed value. The bias table is still
learned, but $\ell$ is masked.
\item[\textsc{Scheduled}.] $g(t)$ follows a manually specified schedule and
$\ell$ is masked as in \textsc{Forced}.
\end{description}
When $\ell$ is masked it remains near its initialization, so normal
evaluation may reactivate a prior that was not used at the end of training.
For this reason, \emph{Scheduled} evaluation uses the final scheduled gate
value and is the appropriate evaluation for scheduled runs. For fade-to-zero,
\emph{Scheduled} and \emph{Zero} are identical. \emph{Normal} evaluation is
kept only as a diagnostic of what happens if the untrained gate logit is
used.

\subsection{Training Paths}
\label{sec:method_path}

The main paths share optimizer, model size, data, and total step budget.
\unforced{} learns both gates freely. \forcedzero{} trains with
$g{=}0$ throughout. \forcedconstHalf{} trains with $g{=}0.5$ throughout.
\linearHalfZero{} decays linearly from $0.5$ at step $0$ to $0$ at the end
of training. The main fade path, \fadeStartSixHundredZero{}, holds
$g{=}0.5$ until step $600$, linearly fades to $0$ over $450$ steps, and
then trains with $g{=}0$ until step $3{,}600$.

Training begins with the prior active, fades it, and then consolidates with
the prior absent. Controls with a hard switch use the same start and end
values but replace the fade with an abrupt jump. Post hoc continuation takes a final
\unforced{} checkpoint and trains it for $2{,}550$ additional
steps at $g{=}0$, matching the zero-gate budget of the fade path.

\subsection{A Continuation View of Removability}
\label{sec:method_theory}

We model the schedule as a local continuation path rather than a new model
class~\cite{allgower2012numerical,mobahi2015link}.
Let $\mathcal{L}(\theta,g)$ be the training loss of the Transformer
parameters $\theta$ when the prior gate is fixed to $g$. If, in a local
neighborhood, a stable minimizer branch $\theta^\star(g)$ satisfies
$\nabla_\theta \mathcal{L}(\theta^\star(g),g)=0$ and has Hessian
$H_g=\nabla^2_{\theta\theta}\mathcal{L}(\theta^\star(g),g)$ with smallest
eigenvalue $\mu_g>0$, then the implicit function theorem gives
\begin{equation}
\frac{d\theta^\star}{dg}
=-H_g^{-1}A_g,\qquad
A_g=\nabla^2_{\theta g}\mathcal{L}(\theta^\star(g),g).
\label{eq:continuation_branch}
\end{equation}
Thus the quantity that determines whether a schedule is ``smooth enough'' is
not whether it is linear, but the branch displacement at each step
\begin{equation}
\Delta_t
\approx |g_{t+1}-g_t|\,\|H_{g_t}^{-1}A_{g_t}\|.
\label{eq:smoothness_condition}
\end{equation}
Tracking is plausible when $\Delta_t$ stays below the basin radius allowed
by the optimizer. A hard switch replaces many small displacements with one
large perturbation. Since
\begin{equation}
\nabla_\theta \mathcal{L}(\theta^\star(g_0),g_1)
= A_{g_0}(g_1-g_0)+O((g_1-g_0)^2),
\label{eq:gradient_shock}
\end{equation}
an abrupt change creates a gradient shock whose direction need not point
toward the prior-free retrieval circuit. The outcome therefore depends on
when the gate changes: a path with two steps or a hard switch can leave the
tracked basin even though it has the same endpoints. Nonlinear schedules
can work when they keep $\Delta_t$ small in regions of high curvature, but
can fail when too much gate change occurs where the branch is
ill-conditioned.

Post hoc continuation fails for a different reason. Free training with a
large gate can converge to a basin $\theta_\text{prior}$ that depends on the
prior and routes the output through the bias term. After setting $g=0$, the
recovery rate of a prior-free circuit depends on the projection of the
zero-gate gradient onto the subspace $\mathcal{C}$ in which the circuit forms:
\begin{equation}
\text{useful recovery signal}
\;\propto\;
\|P_{\mathcal{C}}\nabla_\theta
\mathcal{L}(\theta_\text{prior},0)\|.
\label{eq:posthoc_signal}
\end{equation}
If the prior has already carried the retrieval computation, this projection
can be small or poorly conditioned even when the zero-gate loss is high. The
optimizer then spends its budget moving within the basin formed while the prior was active
instead of entering the prior-free branch. The hard switch and post hoc
experiments test these two failure modes separately.

Linear regression ICL exposes the boundary of this argument. In linear
ICL, \forcedzero{} training already finds a good zero-gate solution, so the
prior-free branch is reachable from random initialization. In discrete
retrieval, by contrast, early zero-gate gradients are sparse and symmetric
until some heads learn where to route value from; the prior supplies a dense
shaping signal that makes the useful branch reachable. The prediction is
therefore conditional, not universal: fade-to-zero should help most when
zero-gate training from scratch cannot enter the retrieval branch, but a
prior-active trajectory can track into it.

\subsection{Models and Tasks}
\label{sec:method_models}

The default Transformer has two layers, four heads,
$d_\text{model}{=}64$, sequence length $32$, vocabulary size $32$, and
about $35$K parameters. A larger model ($L{=}4$, $d{=}128$, about $265$K
parameters) checks that the main effect is not a capacity artifact.

We evaluate associative recall, Markov induction, and linear regression ICL.
Associative recall presents pairs of keys and values and asks for the value
associated with a query key. Markov induction asks for the successor of a
repeated context token under repeat probability
$p\in\{0.2,0.5,0.8\}$. Linear regression ICL samples a random linear
function and predicts the query output from in-context examples. The
trajectory claim focuses on the two discrete retrieval tasks; linear ICL is
used as a boundary case.

%% file: sections/experimental_setup.tex
\section{Experimental Setup}
\label{sec:setup}

\paragraph{Baselines and metrics.}
Static comparisons use \textsc{vanilla}, \textsc{NoPE},
\textsc{RoPE}~\cite{su2024roformer}, \textsc{ALiBi}~\cite{press2022train},
learned RPE~\cite{shaw2018self}, \textsc{DAPE}~\cite{zheng2024dape},
\textsc{CoPE}~\cite{golovneva2024contextual}, \textsc{FIRE}~\cite{li2024functional},
\textsc{Kerple}~\cite{chi2022kerple}, \textsc{GPSA}~\cite{d2021convit},
and three soft-prior variants: \softbias{}, \softpos{}, and
\softcontent{}. All methods use matched backbone dimensions, optimizer, and
training budget. We report accuracy for associative recall and Markov
induction, and mean squared error for linear regression ICL.

\paragraph{Training details.}
Unless otherwise stated, runs use AdamW with learning rate $10^{-3}$,
weight decay $0.01$, gradient clipping at $1.0$, batch size $64$, and
$512$ validation examples per evaluation. We average the main trajectory
results over five seeds for the fade window and causal controls, and over
three seeds for the capacity, shape, task boundary, and mechanism experiments.

\paragraph{Run inventory.}
The experiment archive contains the original trained runs plus $15$ runs
over schedule shapes, $12$ linear trajectory runs, and two mechanism traces
in addition to the original trace for seed 1.

%% file: sections/main_results.tex
\section{Static Baselines}
\label{sec:main_results}

Before the removability tests, the backbone with a soft prior performs well
on all three tasks. On associative recall, \softbias{} reaches
$0.794 \pm 0.022$, the highest mean among thirteen baselines; \softpos{}
is close at $0.783$, while \softcontent{} is near chance ($0.067$). The
useful bias therefore comes from position rather than content. On linear
regression ICL, \softpos{}, learned RPE,
\softbias{}, and \textsc{DAPE} all lie within about $1.5\%$ eval loss.

\begin{figure}[t]
\centering
\begin{subfigure}[t]{0.49\linewidth}
\centering
\includegraphics[width=\linewidth]{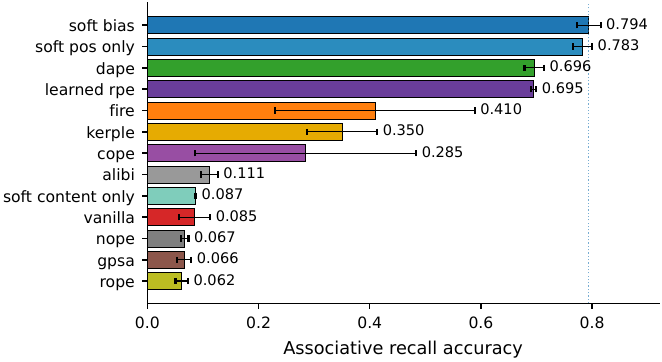}
\caption{Associative recall accuracy.}
\end{subfigure}
\hfill
\begin{subfigure}[t]{0.49\linewidth}
\centering
\includegraphics[width=\linewidth]{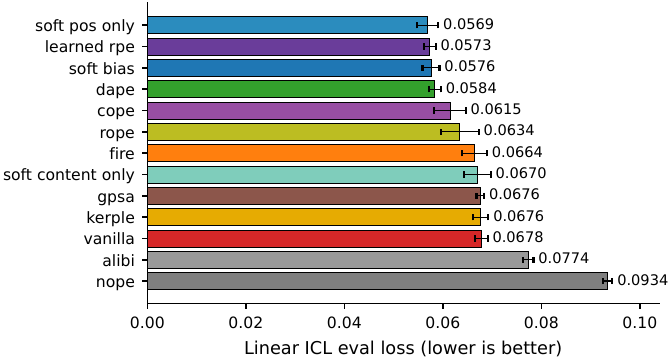}
\caption{Linear ICL eval loss.}
\end{subfigure}
\caption{Static baselines across position and attention variants. The soft
prior performs competitively before any removability test.}
\label{fig:main_static}
\end{figure}

On Markov induction, \softbias{} is also among the strongest methods,
especially at the hardest repeat probability $p{=}0.2$.
These baselines confirm that the soft prior is accurate enough for a
meaningful test of whether the trained model still needs it.

\begin{figure}[t]
\centering
\includegraphics{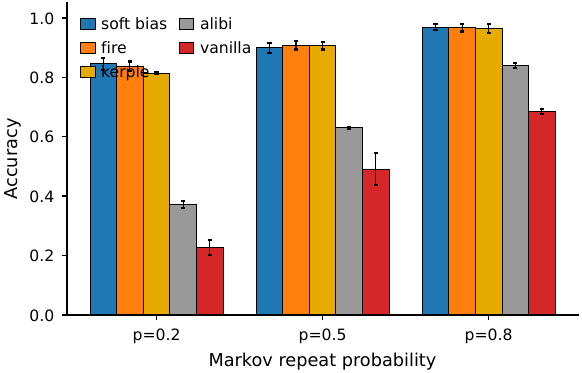}
\caption{Markov induction accuracy across repeat probabilities. Top methods
are close at high repeat probability; the hardest setting separates them.}
\label{fig:markov_sweep}
\end{figure}

%% file: sections/causal_analysis.tex
\section{Gate Interventions}
\label{sec:causal}

\begin{figure}[t]
\centering
\includegraphics{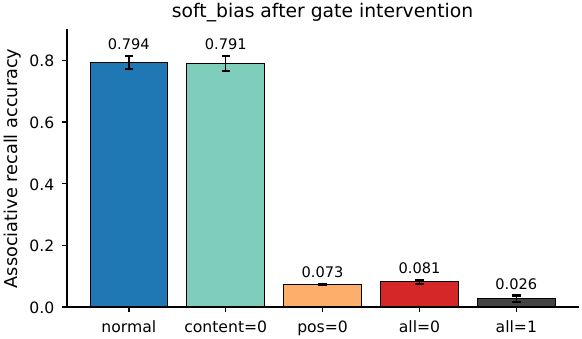}
\caption{Causal gate intervention on associative recall. The unforced model
depends on the positional gate at inference; the content gate is inert on
this task.}
\label{fig:gate_intervention}
\end{figure}

We first test whether the soft prior is causally used by a normally trained
\softbias{} model. At evaluation only, we zero the content gate, the
position gate, or both gates. Figure~\ref{fig:gate_intervention} shows that
content removal is nearly neutral ($0.791$ versus $0.794$ normal accuracy),
whereas position removal collapses accuracy to $0.073$ and removing both
gates gives $0.081$. We retain the content gate only as an architectural
control.

Normal accuracy alone therefore does not establish removability. Under
unforced training, the position prior becomes part of the computation.

%% file: sections/training_path_boundary.tex
\section{Training Path Determines Removability}
\label{sec:training_path}

\subsection{Path Comparison}
\label{sec:path_compare}

\begin{figure}[t]
\centering
\includegraphics[width=\linewidth]{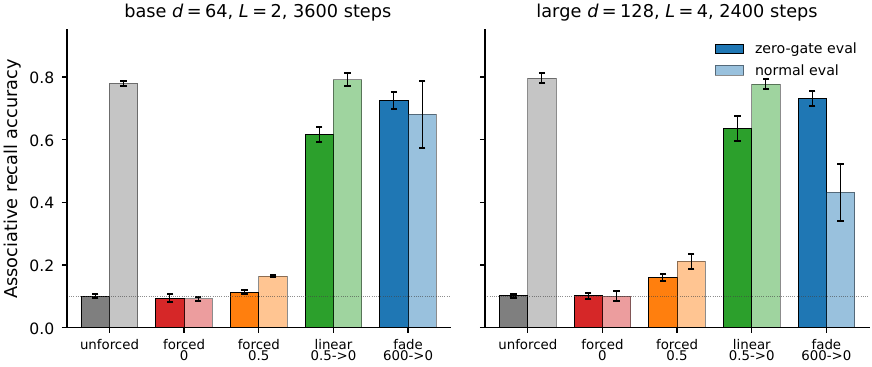}
\caption{Associative recall under matched training paths. For scheduled
runs, zero-gate evaluation uses the gate value reached at the end of training.}
\label{fig:training_path_boundary}
\end{figure}

Figure~\ref{fig:training_path_boundary} and Table~\ref{tab:path_compare}
compare five training paths on associative recall. The unforced model is
strong in normal evaluation but fails when the gate is zeroed. The
\fadeStartSixHundredZero{} path preserves high zero-gate accuracy in both
the base and larger backbones. \forcedzero{} fails on this discrete
retrieval task despite spending the entire run at the final gate value.

\begin{table}[t]
\centering
\caption{Comparison of training paths on associative recall. The base
\fadeStartSixHundredZero{} row reports the estimate over five seeds used
throughout the paper; other rows are matched runs over three seeds.}
\label{tab:path_compare}
\small
\begin{tabular}{lcccc}
\toprule
 & \multicolumn{2}{c}{Base ($L{=}2$, $d{=}64$)} &
\multicolumn{2}{c}{Large ($L{=}4$, $d{=}128$)} \\
\cmidrule(lr){2-3}\cmidrule(lr){4-5}
Path & Zero & Normal & Zero & Normal \\
\midrule
\fadeStartSixHundredZero{} & $.734 \pm .028$ & $.694 \pm .078$ & $.731 \pm .024$ & $.432 \pm .091$ \\
\linearHalfZero{}          & $.617 \pm .024$ & $.792 \pm .022$ & $.635 \pm .040$ & $.778 \pm .015$ \\
\forcedconstHalf{}         & $.114 \pm .006$ & $.164 \pm .003$ & $.160 \pm .011$ & $.212 \pm .024$ \\
\unforced{}                & $.100 \pm .006$ & $.780 \pm .007$ & $.102 \pm .005$ & $.797 \pm .017$ \\
\forcedzero{}              & $.094 \pm .013$ & $.092 \pm .006$ & $.102 \pm .010$ & $.100 \pm .016$ \\
\bottomrule
\end{tabular}
\end{table}

\subsection{Fade Window and Schedule Shape}
\label{sec:trajectory_sweep}

\begin{figure}[t]
\centering
\includegraphics{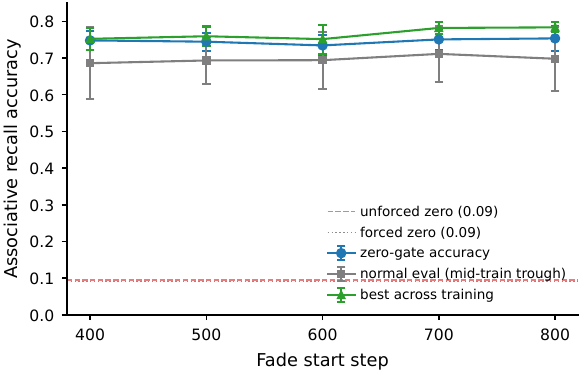}
\caption{Sweep over fade start times with five seeds. Zero-gate accuracy is stable
across steps $400$--$800$.}
\label{fig:trajectory_sweep}
\end{figure}

The effect is stable over a broad range of fade start times. Across five seeds,
zero-gate accuracy is $0.748$, $0.745$, $0.734$, $0.751$, and $0.754$ for
fade starts $400$, $500$, $600$, $700$, and $800$, respectively
(Fig.~\ref{fig:trajectory_sweep}). In contrast, the unforced and forced-zero
baselines remain near $0.09$.

Smoothness also matters. Holding the same start value, end value, fade start,
and fade width, linear, cosine, and convex schedules all work
($0.749$--$0.764$ zero-gate accuracy), while a schedule with two steps reaches
only $0.212$. A concave schedule is unstable across seeds. Thus the result
is not specific to a linear formula, but it does require a gradual path
rather than an abrupt deletion.

\begin{figure}[t]
\centering
\includegraphics{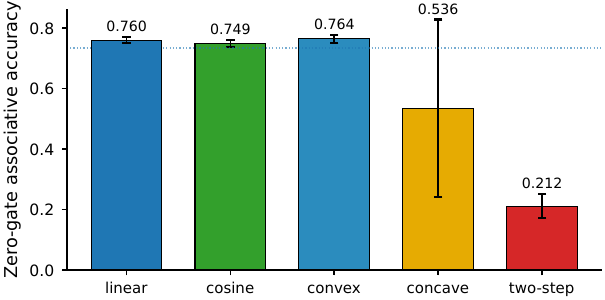}
\caption{Scan over schedule shapes on associative recall (3 seeds). Smooth
fades work; a deletion in two steps does not.}
\label{fig:schedule_shapes}
\end{figure}

\subsection{Causal Controls}
\label{sec:causal_controls}

Two controls rule out simpler explanations. A hard switch asks whether the
model only needs a warm start with an active prior followed by zero-gate
training. Post hoc continuation asks whether the fade result comes merely
from extra zero-gate training. Table~\ref{tab:causal_controls} shows that neither is
sufficient. Switching abruptly at the fade endpoint reaches only
$0.293 \pm 0.128$, and continuation from an unforced checkpoint reaches
$0.178 \pm 0.035$.

\begin{table}[t]
\centering
\caption{Causal controls on associative recall (5 seeds).}
\label{tab:causal_controls}
\small
\begin{tabular}{lcc}
\toprule
Condition & Zero-gate acc & Normal acc \\
\midrule
\fadeStartSixHundredZero{} & $.734 \pm .028$ & $.694 \pm .078$ \\
Hard switch at step $1050$ & $.293 \pm .128$ & $.367 \pm .139$ \\
Hard switch at step $600$  & $.097 \pm .020$ & $.102 \pm .024$ \\
Post hoc zero continuation & $.178 \pm .035$ & $.619 \pm .026$ \\
\unforced{}                & $.095 \pm .009$ & $.772 \pm .020$ \\
\forcedzero{}              & $.091 \pm .012$ & $.088 \pm .010$ \\
\bottomrule
\end{tabular}
\end{table}

\begin{figure}[t]
\centering
\includegraphics[width=\linewidth]{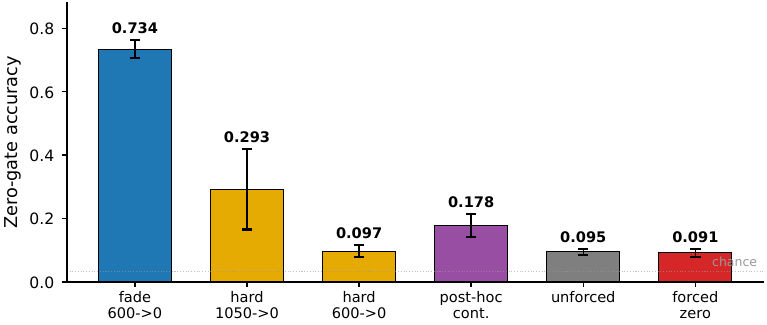}
\caption{Hard switches and post hoc continuation fall far below smooth
fade-to-zero training.}
\label{fig:reviewer_controls}
\end{figure}

These failures locate the effect in the training path: the prior must shape
the circuit while it forms, and the gate must decrease gradually enough for
the internal attention pattern to take over.

%% file: sections/cross_task.tex
\section{Boundary Across Tasks}
\label{sec:cross_task}

Markov induction replicates the pattern from discrete retrieval. At repeat
probabilities $p{=}0.2$, $0.5$, and $0.8$, \fadeStartSixHundredZero{}
retains $0.830$, $0.920$, and $0.969$ zero-gate accuracy, respectively.
Unforced training collapses much more strongly, especially at the hard
$p{=}0.2$ setting.

\begin{figure}[t]
\centering
\includegraphics[width=\linewidth]{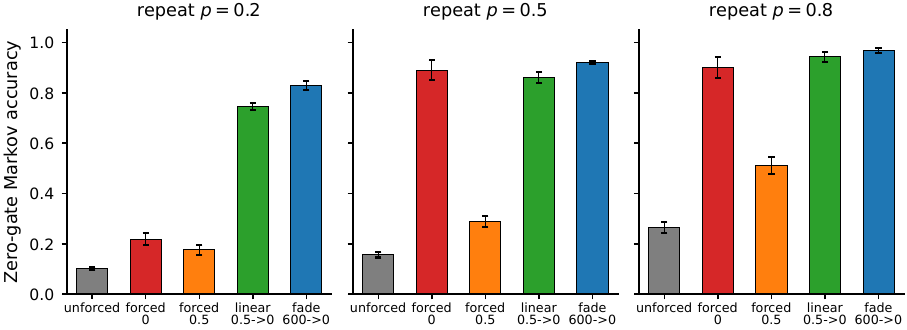}
\caption{Markov induction zero-gate accuracy across training paths and
repeat probabilities. Fade-to-zero remains strong; unforced training is not
removable.}
\label{fig:markov_boundary}
\end{figure}

Linear regression ICL behaves differently. It still shows a removability problem
for unforced training: normal loss is $0.0656$, while zero-gate loss degrades
to $0.1296$. However, unlike associative recall, \forcedzero{} learns a
reasonable solution directly ($0.0722$ loss), and \fadeStartSixHundredZero{}
is only slightly better ($0.0646$). This result bounds the claim: gradual
fading helps most on discrete retrieval circuits for which zero-gate
training from scratch fails. The fade advantage does not hold for every ICL task.

\begin{figure}[t]
\centering
\includegraphics{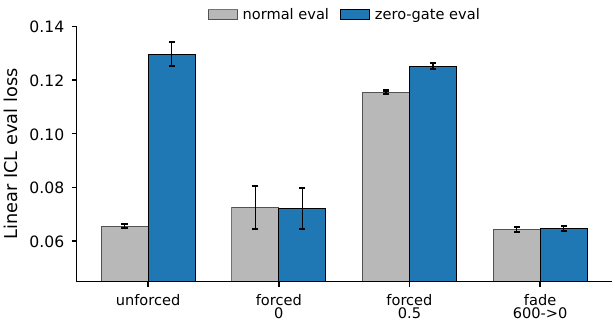}
\caption{Linear regression ICL trajectory test (3 seeds). Unforced training
is not removable, but forced-zero training can solve the task, making this a
boundary case rather than a positive replication of the discrete retrieval
effect.}
\label{fig:linear_trajectory}
\end{figure}

\paragraph{Length generalisation.}
Associative recall accuracy degrades when evaluated far beyond the training
length. At $T{=}120$, \softbias{} reaches only $0.263$. We therefore do not
claim that removability improves length extrapolation; the length result is
a limitation of the recovered prior-free circuit.

%% file: sections/mechanistic.tex
\section{Mechanistic Diagnostics}
\label{sec:mech}

We trace three \fadeStartSixHundredZero{} seeds every $300$ steps. At each
snapshot we measure (i) successor-value mass, the attention-value mass a
head places on the correct successor token, and (ii) zero-gate ablation
importance, the accuracy drop when that head is replaced by its batch mean.

\begin{table}[t]
\centering
\caption{Mechanism trace summary across seeds. Head identity varies, but
the timing of consolidation is stable.}
\label{tab:mechanism_multiseed}
\small
\setlength{\tabcolsep}{1pt}
\begin{tabular}{ccccc}
\toprule
Seed & Top successor head & First successor $\ge .25$ &
Top ablation head & First ablation $\ge .03$ \\
\midrule
0 & L1H0 & 1500 & L0H2 & 1200 \\
1 & L1H3 & 1500 & L0H0 & 1200 \\
2 & L1H1 & 1500 & L0H3 & 1200 \\
\bottomrule
\end{tabular}
\end{table}

The head carrying successor mass changes by seed, as expected under head
permutation symmetry (Table~\ref{tab:mechanism_multiseed}). The timing does
not: ablation importance first crosses
the threshold at step $1{,}200$, and successor-value mass first crosses the
threshold at step $1{,}500$ in all three seeds. Both occur after the fade
ends around step $1{,}050$.

\begin{figure}[!b]
\centering
\includegraphics[width=0.92\linewidth]{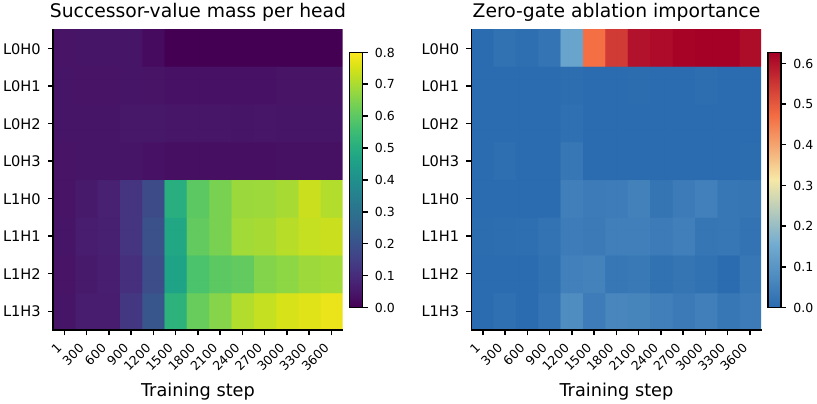}
\caption{Head evolution for representative seed 1. Successor-value mass (left)
separates functional routing from zero-gate ablation importance (right). The
strongest successor and ablation signals appear after the gate has reached
zero, supporting delayed consolidation rather than a universal head
identity.}
\label{fig:head_evolution}
\end{figure}

Figure~\ref{fig:head_evolution} identifies the timing, not a universal head
identity. For this representative seed,
$L0H0$ becomes critical at zero gate after step $1{,}200$, while $L1H3$ carries
the largest successor-value mass after step $1{,}500$. This delayed
consolidation matches the hard switch and post hoc controls in
Section~\ref{sec:causal_controls}.

%% file: sections/limitations.tex
\section{Limitations and Conclusion}
\label{sec:limit}

All positive trajectory results come from synthetic discrete retrieval tasks
with Transformers no larger than $L{=}4$ and $d{=}128$. Linear regression
ICL does not require the same formation path aided by the prior, and length extrapolation
remains weak. The content gate is mostly inert in the tested tasks, so the
current evidence concerns positional priors rather than content priors. The
scan over schedule shapes is also limited to a few simple curves; it supports
smoothness over abrupt deletion, not an exhaustive theory of all possible
annealing paths.

Under unforced training, a soft positional prior can become a permanent
dependency. Under a fade-to-zero path, the same prior can guide training and
then be removed. Hard switching and post hoc continuation show that neither
a warm start nor extra zero-gate training explains the difference. The
mechanism traces place consolidation after the prior is removed even though
head identities vary by seed. For small discrete retrieval circuits,
removability therefore depends on the training trajectory.